\documentclass[conference]{IEEEtran}
\usepackage[T1]{fontenc}
\usepackage[utf8]{inputenc}
\usepackage{fancyvrb}
\usepackage{graphicx}
\usepackage{listings}
\usepackage{multirow} 
\usepackage{multicol} 
\usepackage{float} 
\usepackage{listings}
\usepackage{xcolor}
\usepackage{amsfonts}
\usepackage{amsmath}
\usepackage{tikz}
\usetikzlibrary{positioning}
\usepackage{xcolor}
\usepackage[normalem]{ulem}
\PassOptionsToPackage{hyphens}{url}
\usepackage{hyperref} 
\usepackage{rotating}

\makeatletter % changes the catcode of @ to 11
\newcommand{\linebreakand}{%
  \end{@IEEEauthorhalign}
  \hfill\mbox{}\par
  \mbox{}\hfill\begin{@IEEEauthorhalign}
}
\makeatother % changes the catcode of @ back to 12
\newcommand{\probP}{\text{I\kern-0.15em P}}
\def\BibTeX{{\rm B\kern-.05em{\sc i\kern-.025em b}\kern-.08em
    T\kern-.1667em\lower.7ex\hbox{E}\kern-.125emX}}

\usepackage{float}

\lstdefinestyle{pythonstyle}{
    language=Python, % Langage Python
    basicstyle=\ttfamily\small, % Style de base : petite taille, police monospace
    keywordstyle=\color{blue}\bfseries, % Mots-clés en bleu gras
    commentstyle=\color{green!50!black}\itshape, % Commentaires en vert italique
    stringstyle=\color{red}, % Chaînes de caractères en rouge
    backgroundcolor=\color{gray!10}, % Fond gris clair
    frame=single, % Cadre autour du code
    rulecolor=\color{gray}, % Couleur du cadre
    tabsize=4, % Taille des tabulations
    showstringspaces=false, % Ne pas montrer les espaces dans les chaînes
    breaklines=true, % Permet la coupure des lignes longues
    breakatwhitespace=true, % Coupure aux espaces
    numbers=none, % Pas de numéros de ligne
}

\begin{document}

\title{Khiops: An End-to-End, Frugal AutoML and XAI\\ Machine Learning Solution\\ for Large, Multi-Table Databases}

\author{Marc Boullé,
        Nicolas Voisine, Bruno Guerraz, Carine Hue, Felipe Olmos, Vladimir Popescu, Stéphane Gouache,\\
        Stéphane Bouget,        Alexis Bondu, Luc Aurelien Gauthier,  Yassine Nair Benrekia, 
        Fabrice Clérot, Vincent Lemaire\\
        (Orange Research,  contact: firstname.name@orange.com)}

\maketitle

\begin{abstract}%\textcolor{magenta}{remplacer tool par library partout ?  }
Khiops is an open source machine learning tool designed for mining large multi-table databases. 
Khiops is based on a unique Bayesian approach that has attracted academic interest with more than 20 publications on topics such as variable selection, classification, decision trees and co-clustering. It provides a predictive measure of variable importance using discretisation models for numerical data and value clustering for categorical data. The proposed classification/regression model is a naive Bayesian classifier incorporating variable selection and weight learning. In the case of multi-table databases, it provides propositionalisation by automatically constructing aggregates.
Khiops is adapted to the analysis of large databases with millions of individuals, tens of thousands of variables and hundreds of millions of records in secondary tables. It is available on many environments, both from a Python library and via a user interface.
\end{abstract}


\section{What makes Khiops different}

\begin{figure*}[!htb]
\begin{center}
 \includegraphics[width=1.0\textwidth]{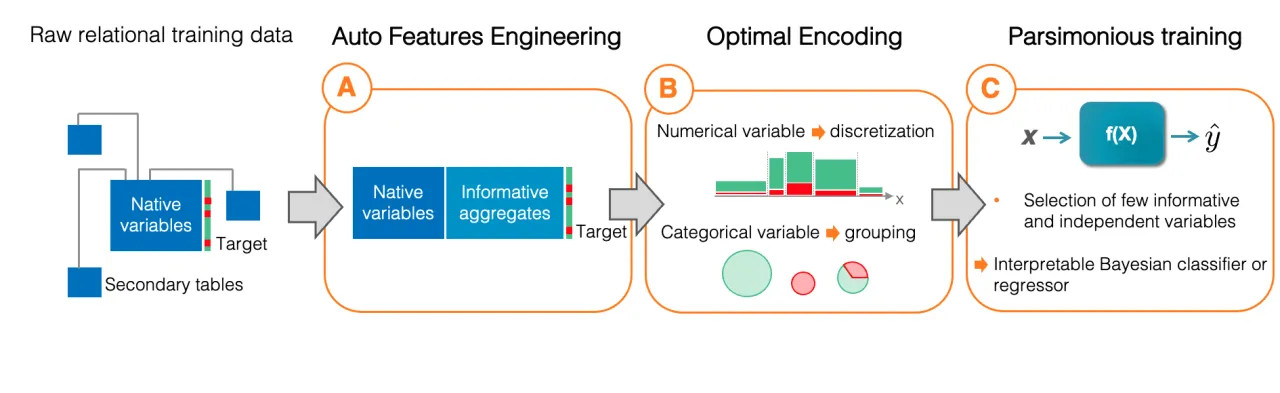}
 \caption{Machine learning process implemented by Khiops}
 \label{fig_chaine}
\end{center}
\end{figure*}

Khiops is an end-to-end Machine Learning (AutoML) solution that natively and effortlessly handles complex and time-consuming data science tasks on multi-million instance datasets. Khiops tasks include variable engineering (A), data cleaning and encoding (B), and parsimonious model learning (C) (see Figure \ref{fig_chaine}).  Khiops also includes features that allow it to be fully explainable (XAI).

The AutoML capability allows Khiops to process tabular or relational data with complex star or snowflake schemas. This is a real differentiator in a variety of situations, particularly when dealing with use cases with multiple records per statistical individual (such as calls, transactions or production logs). In a world of increasingly sophisticated cyber attacks, log analysis has become a necessity. Imagine being able to identify an intrusion in real time or precisely retrace an attacker's route to limit the damage. That's exactly what effective log management can do \cite{ehis2023optimization,zulfadhilah2016cyber}.

The uniqueness of Khiops lies in its different approach to typical AutoML solutions, which often run an expensive range of complex algorithms on parameter sets using grid search. Instead, Khiops uses an original formalism called MODL (which is hyperparameter-free), allowing it to push the boundaries of automation on very large multi-table datasets and push the boundaries of automation. This allows it to build high-performance models that are simple to deploy and easy to interpret. Khiops comes with a low-code Python library that offers an efficient AutoML pipeline in a simple $.fit()$ function. Its sophisticated algorithms are easy to use, thanks to its Python library that follows Scikit-learn (sklearn) standards. Khiops facilitates automatic learning in a complete safety environment. This approach significantly reduces the time spent on the modelling phase, allowing users to allocate more time to analyse their models and gain a deeper understanding of their data, while requiring minimal coding.

Khiops is equipped with an interactive visualisation tool that provides full access to the preparation and modelling results directly from a notebook or dedicated application. Consequently, there is no requirement to write specific visualisation code to present and interpret modelling results. In addition, Khiops offers a version with a graphical interface that allows all learning algorithms to be used without the need to write a single line of code, making it easily usable by business domain specialists without requiring in-depth knowledge of data analysis.

\section{An original Bayesian formalism}

Whether for variable creation, transformation and selection, co-clustering or decision trees, Khiops uses an original Bayesian formalism, MODL \cite{boulle:pastel-00003023}. The MODL approach aims to select the most likely model given the training data. Bayes' formula is therefore the starting point for deriving the optimisation criteria used, the general form of which is as follows:

$$
arg \, \underset{h\in\mathcal{H}}{max} P(h|d) = arg \, \underset{h\in\mathcal{H}}{max} \frac{P(h)P(d|h)}{P(d)}
$$

All MODL optimisation criteria are designed in the same way (optimal coding, automatic variable engineering and parsimonious learning), according to the following steps:

\begin{itemize}
 \item define the $\mathcal{H}$ family of models, i.e. the modelling parameters, as a function of the learning task to be performed (i.e. $\mathcal{H}$ can be a discretization \cite{BoulleML06}, a grouping of values \cite{BoulleJMLR05} or a decision tree \cite{voisineEtAlAKDM09});
 \item define the prior distribution on these parameters $P(h)$, which is always hierarchical and uniform at each stage of the hierarchy;
\item obtain an optimisation criterion from the development of Bayes' formula, taking into account the likelihood term $P(d|h)$;
 \item learn the model by optimising the final criterion.
\end{itemize}

In information theory, the model selection problem described above can be translated into an encoding problem, the aim of which is to find the most compact way of encoding an information source for transmission over a telecommunications channel. Consider an information source emitting symbols [for example, a, b, c, etc.] whose alphabet is known. In information theory, the negative logarithm of the probability of a symbol being transmitted ($-log(P(a))$) represents its optimal coding length, denoted by $L$ and expressed in bits. According to Shannon's intuition, the most efficient encoding strategy is to assign a short coding length to the most frequent symbols. Similarly, the probabilities in Bayes' formula above can be replaced by negative logarithms to obtain a MODL criterion to minimise, which can be interpreted as follows:

$$
-log(P(h).P(d|h)) = \underbrace{L(h)}_{Prior}+\underbrace{L(d|h)}_{Likelihood}
$$

\begin{itemize}
\item the prior corresponds to the coding length of the model, i.e. the number of bits needed to describe it;
 \item the likelihood is the coding length of the training data knowing the model.
\end{itemize}

In this particular instance of the encoding problem, the model is first transmitted over the telecommunications channel, followed by the data. The Minimum Description Length (MDL) principle aims to select the most compact model describing the data, and is applied in the MODL approach by choosing a hierarchical prior representing successive choices of model parameters.

\section{Tool presentation}

The Khiops tool integrates the work carried out at Orange Research on data preparation, automatic construction of variables for multi-table databases and large scale modelling.
Since 2024, the Khiops V10 version has been open source.
The very recent last version (V11) includes the following main features:

\begin{itemize}
	\item management of multi-table data, 
	\item automatic feature construction to generate a flat table of individuals $\times$ variables, 
    \item automatic feature construction from text variables,
	\item optimal data preparation via discretisation and value grouping,
    \item random forests for classification and regression,
	\item modelling using a naive Bayesian classifier, with optimal univariate pre-processing, variable selection and learning of weights for each variable,
	\item deployment of models directly on multi-table bases,
    \item interpretation and reinforcement models,
    \item optimal histograms for univariate data exploration,
    \item variable $\times$ variable coclustering, for joint density estimation,
    \item instance $\times$ variable coclustering, for exploratory analysis,
    \item end-to-end management of sparse data,
    \item handling of local as well as cloud storage.\\
\end{itemize}

The tool is written in C++ for the algorithmic component and Java for the graphical interface. It can be used with either a graphical user interface or a Python library, allowing for easy integration into a processing pipeline.. There is also an interactive visualisation tool available for inspecting the results of preparation, modelling and evaluation (see Figure \ref{fig_visu}).\\

Khiops is available at \url{http://www.khiops.org}. The current version (V11) is used in a wide range of applications: including customer marketing (attrition models, appetite for new services, etc.), text mining, web mining, banking, social networks, technical and economic studies, internet traffic characterisation, ergonomics, user sociology, fraud detection, ... It has been used with learning databases containing millions of individuals and hundreds of millions of secondary records.\\

\subsection{Installation :}
The Khiops python library is easy to install using the conda package manager.

\begin{lstlisting}[language=bash]
# Windows/Linux/macOS
conda install khiops -c conda-forge -c khiops
\end{lstlisting}

\subsection{ Automatic variable construction:}

In the case of multi-tables, this is one of the major contributions of the tool. It is based on the description of a multi-table star or snowflake schema\footnote{The terminology used is similar to that of data warehouses, such as star or snowflake schemas. However, here we are not talking about concepts for structuring a data warehouse, but rather about describing individuals in a statistical analysis, with some variables coming from the root table and others from secondary tables.}, with a root table containing the individuals to be analysed (e.g. customers) and secondary tables in 0-1 or 0-n relationships containing records completing the description of the individuals (e.g. communication details).

The only user parameter is then the number of variables to be constructed, by systematically applying selection or aggregation functions. This method used \cite{boulle2019scalable} exploits a Bayesian regularisation approach based on a parsimonious prior distribution over the potentially infinite set of all the variables that can be constructed. Variables are then constructed using an efficient sampling algorithm according to this prior distribution. The resulting method is simple to use, computationally efficient and robust to the problem of overlearning. The creation of MODL decision trees is the final step in the AutoML pipeline implemented by Khiops. This  optional pre-processing step involves building decision trees from native variables and aggregates \cite{voisineEtAlAKDM09}, resulting in the model as a parsimonious AutoML "random forest" model. \\

The important point to understand here is that the users only need to provide the schema of their data and the number of variables to be constructed, with selection and aggregation functions applied systematically. The process is fully automated. This saves a great deal of time, as there is no need to build aggregates by hand, which would require considerable business expertise. It also means that aggregates can be discovered on subjects where the Khiops user is not an expert, as well as new aggregates (new knowledge) on familiar subjects. The ‘accident’ use case below is clearly not an example of cyber defence, but all the principles remain valid.\\

\subsection{Optimal preparation :}
Data is prepared using supervised discretisation \cite{BoulleML06} for numerical variables and supervised grouping of values \cite{BoulleJMLR05} for categorical variables. The associated methods exploit a Bayesian model selection approach to construct the most likely preparation model given the data, which provides an accurate and robust estimate of the univariate conditional density per descriptive variable.\\

\subsection{Parsimonious learning:} 
Modelling takes advantage of all initial variables, as well as those constructed after preparation, combining them using a naïve Bayesian classifier with variable selection and direct learning of weights per variable \cite{hue2024fractionalnaivebayesfnb}.\\

\subsection{Automatic adaptation to material resources:}
Khiops adapts its algorithms to the available hardware resources  (RAM and CPU). Khiops divides the data into a more or less fine-grained matrix of files by partitioning the instances into rows and the variables into columns, depending on the learning task in hand and the hardware resources available. The successive stages of the AutoML pipeline are algorithms that process either rows or columns of the root table. For example, optimal encoding is a column-based algorithm, since each discretisation or clustering model can be learned independently for each variable. On the other hand, once the pipeline is executed, making predictions is a row-based algorithm, since each example can be processed independently. The aim is to optimise the execution time of these algorithms, whatever the size of the data processed and the amount of hardware resources available. Take, for example, the Zeta classification problem (9.3 GB) of the Large Scale Learning Challenge \cite{SonnenburgEtAl08}, which contains $500,000$ training examples and $2,000$ numerical explanatory variables. Learning on an Intel Xeon Gold 6150 2.70 Ghz processor takes 81 minutes with a single core and 512 MB of RAM, and only 3 minutes with 32 cores and 16 GB of RAM (See Section~\ref{zeta_example} for more details on the Zeta problem).\\

{\subsection{Interfaces :}
Although Khiops provides a core Python library \texttt{\small{khiops.core}} to effectively meet the challenge of large volumes, it is also possible to start with the \texttt{\small{khiops.sklearn}} library for those familiar with the popular sklearn library, or even to use a GUI with Khiops Desktop. Online deployment of Khiops models for real-time applications can be done using the KNI library. Finally, it should be noted that models learned by Khiops can be easily interpreted using dedicated visualization tools.

\subsection{Khiops is an environmentally-friendly tool (frugal \cite{10.1145/3748239.3748247}):}
The Khiops code is highly optimised: (i) advanced optimisation algorithms have been designed specifically for each type of learning task, (ii) they have been implemented in “low-level” C++ using very fine-grained optimisation close to the hardware layer. Khiops intelligently adapts the execution of algorithms to the available hardware resources, taking into account the size of the task to be executed. The solution is compact enough to be embedded. In this way, Khiops is able to run transparently on a Rasberry, a phone, ... , with data that far exceeds the available RAM, or on a Kubernetes cluster by adapting the number of nodes used to the size of the data. There is never any need to invest in large hardware, as execution time is the adjustment variable: ‘Khiops does the best in all cases’. The models generated by Khiops adapt to the data and the machine learning task. For a simple problem, Khiops produces a parsimonious, intelligible model with few parameters, and therefore inexpensive to deploy and interpret! Khiops uses data reduction natively (parsimony): the model explicitly selects a subset of the variables and only these variables are required for deployment.

\subsection{Khiops is an XAI tool :\label{xai2}}   As described below in the example on the 'Accidents' database Khiops also offers an interactive results visualization tool, called Khiops Visualization (figure \ref{fig_visu}). This tool allows to visualize all analysis results in an intuitive way, offering a quick and easy interpretation. This visualisation tool allow to interpret the model’s global behaviour for the whole dataset. But the tool also offer the possibility to obtain local behaviour, local explanations per example. Firstly by computing the Shapley values of all the input variables of a trained classifier for each example of a deployment (or test) dataset, see \cite{10.1007/978-3-031-74630-7_6} for more details. Secondly by suggesting variable change (univariate change) to improve (to reinforce) the probability to belong to a class on interest (in the sense of a counterfactual but where the value\footnote{The computation of a `complete counterfactual' will be available in 2025 on \url{www.khiops.org} as a notebook python \cite{lemaire2024viewing}.} of a single variable has been changed) see \cite{LemaireKDD2009correlation} (Section 4) for more details.

\section{Examples of use}

\subsection{The Accident Database}

In this example, we will show how Khiops can be used to train a classifier on complex relational data where a secondary table is itself a parent table of another table (i.e. a flake schema). We will train a multi-table classifier on the Accidents dataset. The Accidents database lists all the accidents involving injuries that occurred during 2018 in France, with a simplified description.\\

This database includes the following information:
\begin{itemize}
\item The location of the accident (Places table);
\item The characteristics of the accident (Accidents table);
\item The vehicles involved (Vehicles table);
\item The passengers in the vehicles (Users table);
\end{itemize}

The data is organised according to the following relational snowflake schema.

\begin{verbatim}
Accidents
|
| -- 1:n -- Vehicles
|             |
|             |-- 1:n -- Users
|
| -- 1:1 -- Places
\end{verbatim}

To train the \texttt{KhiopsClassifier} with this data, we then need to specify a multi-table dataset: the main table \textbf{Accidents}, the secondary tables \textbf{Vehicles} and \textbf{Places}, the tertiary table \textbf{Users}.\\

\subsubsection{Multi-table specification:}
The first step is to specify the schema of the multi-table dataset. Khiops offers an extension to sklearn's single-table description. The main Accidents table and the secondary Places table have a single key: `AccidentId'. The Vehicles (the secondary table) and Users (the tertiary table) tables have a key with two fields: `AccidentId' and `VehicleId'. To describe the relationships between the tables, the relationships field must be added to the table specification dictionary. For a $0:1$ relationship instead of $0:n$, `True' must be added at the end of the relationship specification (see Figure \ref{list_python1}):

\begin{figure}[H] % H pour forcer la position
    \centering
    \begin{lstlisting}[style=pythonstyle,basicstyle=\footnotesize]
X_accidents_train = {
    "main_table": (accidents_df.drop("Gravity", axis=1), ["AccidentId"]),
    "additional_data_tables": {
        "Vehicles": (vehicles_df, ["AccidentId", "VehicleId"]),
        "Vehicles/Users": (users_df, ["AccidentId", "VehicleId"]),
        "Places": (places_df, ["AccidentId"], True),
    },
}
y_accidents_train = accidents_df["Gravity"]
    \end{lstlisting}
    \caption{Specification of the multi-table dataset}
    \label{list_python1}
\end{figure}

\begin{table*}[!t]
\setlength{\tabcolsep}{1pt}
\centering
\fontsize{6.5}{7}\selectfont
\begin{tabular}{|c|l|c|c|l|c|c|}
\hline
ProbGravityLethal	& ShapleyVariable\_Lethal\_1	             & ShapleyPart\_Lethal\_1	   & ShapleyValue\_Lethal\_1   & ShapleyVariable\_Lethal\_2	         & ShapleyPart\_Lethal\_2	  &  ShapleyValue\_Lethal\_2\\ \hline
0,708149757	       &  Max(Vehicles,Min(Users,BirthYear))	    &  ]-inf,1933,5]	           & 0,468556017	           &  Min(Vehicles,Min(Users,BirthYear))	 &    ]-inf,1933,5]	      &  0,407707682\\ \hline
0,688394752	       &  Max(Vehicles,Min(Users,BirthYear))	    &  ]-inf,1933,5]	           & 0,468556017	           &  Light	                             & {NightNoStreetLight}	  &  0,419425784\\ \hline
0,575788413	      &   Light	                                &   {NightNoStreetLight}	   & 0,419425784	           &  InAgglomeration	                     &    {No}	              &  0,321445857\\ \hline
0,548183385	      &   Mean(Vehicles,Min(Users,BirthYear))	    & ]-inf,1938,25]	           &  0,363729716	           &  InAgglomeration	                     &      {No}	              &  0,321445857\\ \hline
0,547824738	      &   Light	                                &   {NightNoStreetLight}	   & 0,419425784	           &  Mean(Vehicles,Min(Users,BirthYear))	 &        ]-inf,1938,25]	 & 0,363729716\\ \hline
\end{tabular}
\vspace{0.2cm}
\caption{Illustration of one XAI output that can be provided by Khiops.}
\label{shapley}
\end{table*}

\subsubsection{Learning: }
\label{sec:training}
Like a \texttt{sklearn} classification, it is simply a matter of using the functions \texttt{khc.fit} for learning and \texttt{khc.predict} for deployment (see Figure \ref{list_python2}). In the table \ref{tab:khiops_performance}, we varied $n\_features$ and $max\_cores$ to observe their influence on performance in time and AUC. We quickly noticed that increasing the number of aggregates improved performance, and that increasing the number of cores used greatly reduced analysis time.

\begin{figure}[!h] % H pour forcer la position
    \centering
    \begin{lstlisting}[style=pythonstyle,basicstyle=\footnotesize]
    # Creating a Khiops model with AUTO Feature Multi-table
    khc = KhiopsClassifier(n_trees=0,n_features=10,max_cores=1)
    # Train the model
    khc.fit(X_accidents_train, y_accidents_train)
    # Predict labels
    y_pred = khc.predict(X_accidents_train)
    # Calculate probabilities
    y_probas = khc.predict_proba(X_accidents_train)
    \end{lstlisting}
    \caption{Learning and deploying on the Accidents database}
    \label{list_python2}
\end{figure}

\begin{table}[!ht]
    \centering
    \fontsize{8pt}{8pt}
    \selectfont
    \begin{tabular}{|l|l|l|l|l|l|}
    \hline
        Features number& 10 & 100 & 1 000 & 10 000 & 100 000 \\ \hline\hline 
        Train AUC & 0.792 & 0.826 & 0.845 & 0.865  & 0.874\\ \hline
        Test AUC & 0.781 & 0.818 & 0.838 & 0.855 & 0.854\\ \hline
        Time with 1 core& 3 & 8 & 33 & 273 &2552 \\ \hline
         Time with 5 cores & 3 & 4 & 12 & 76  & 712\\ \hline
         Time with 9 cores & 3 & 4 & 8 & 52 & 438  \\ \hline
    \end{tabular}
    \vspace{0.2cm}
    \caption{Khiops learning performance on the Accidents table according to the number of aggregates generated. Performances include AUC in train and in test, as well as learning time in seconds for 1, 5, and 9 cores.}
    \label{tab:khiops_performance}
\end{table}

\subsubsection{Viewing results:}
Although the core api \texttt{khiops.core} contains all the methods to analyze Khiops results, Khiops also offers an interactive results visualization tool, called Khiops Visualization (figure \ref{fig_visu}). This tool allows to visualize all analysis results in an intuitive way, offering a quick and easy interpretation. \\

Khiops Visualization is composed of several panels. Depending on the analysis type, the panels and their contents are not the same. In case of a supervised analysis (as for the Accident database), Khiops Visualization can be composed with 5 panels (see the top of the figure \ref{fig_visu}: (i) Preparation: displays the Preparation report; (ii) Tree preparation : displays the preparation report for tree variables; (iii) Preparation 2D: displays the 2D preparation report (iv) Modelling: displays the modelling report; (v) Evaluation: displays on one panel the test, train and evaluation reports. Finally Project Infos : displays the report file and database locations plus some short comments on the analysis. All the panels are described in a lot of details on \url{https://khiops.org/ui-docs/visualization/}\\

\begin{figure*}[!htb]
\begin{center}
 \includegraphics[width=0.95\textwidth]{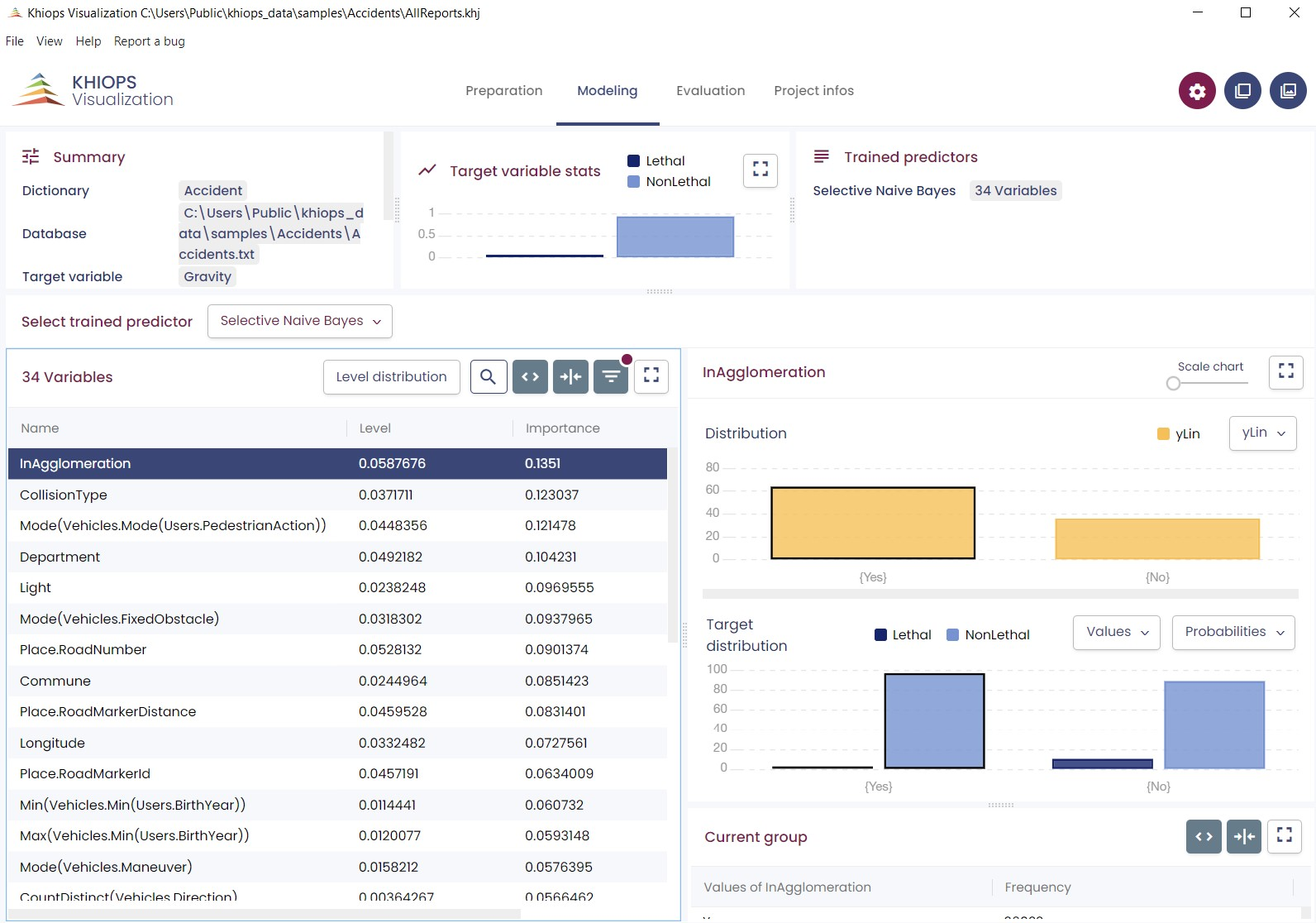}
 \caption{Screenshot of Khiops Visualisation after analysing the accident database and constructing 100 aggregates}
 \label{fig_visu}
\end{center}
\end{figure*}

\subsubsection{Variable Importance results:}
\label{sec:var-importance}

To illustrate one of the XAI aspects (see section \ref{xai2}) of Khiops, we give in Table~\ref{shapley} one  example of the outputs it can provides. This table gives the 5 accidents among the ones with  high probably of being lethal (predicted by the classifier) in the first column.
We ask the tool to give for each accident the two variables which contribute the more to the predicted probability (the number of variables is just define per user when asking this XAI outcome) to be lethal. Therefore here, after the first column, there are 2 triplets of columns. Each triplet gives for each accident the name of the variable, then the value of the variable and finally the Shapley value for this variable. The triplets (so the columns of the file) are sorting according to the Shapley value allowing a fast understanding of the individual variable importances.

When examining the second accident (line 2) in this table, we see that the most important variable is ``Max(Vehicles,Min(Users,BirthYear))'' and the second one is ``Light''.
The value of the most important belongs to the value interval ]-inf,1993.5] while the value of the second most important value belongs to the to the categorical value ``NightNoStreetLight''. The associated shapley values are in columns 4 and 7.
For this accident, the main causes of a high probability of being lethal are therefore easy to understand one of the vehicles involved in the accident involves an old=occupant, born before 1993 and the absence of light in the street during the night. The others lines of this table appear to be equally straightforward to read. \\

Of course Khiops can also output a file with all the Shapley values for all variables and for all the classes, allowing the use of this file with a python library like Shap \cite{NIPS2017_7062} to create personalized visualisation.\\

%%%%%%%%%%%%%%%%%%%%%%%%%%%%%%%%%%%%%%%%%
\subsection{The UNSW-NB15 dataset}

In this section we follow exactly the same process than in the previous section except that we use the Khiops library on the UNSW-NB15 dataset\footnote{\url{https://www.kaggle.com/datasets/mrwellsdavid/unsw-nb15/data}} which is a flat dataset\footnote{In the description of this dataset, it appears to be based on an initial star schema, but it no longer seems to be available. We have contacted the creators of the dataset to request the relational version, but we have not received a response.}.

This UNSW-NB 15 dataset was created by the IXIA PerfectStorm tool in the Cyber Range Lab of the Australian Centre for Cyber Security (ACCS) for generating a hybrid of real modern normal activities and synthetic contemporary attack behaviours. This dataset includes nine types of attacks, namely, Fuzzers, Analysis, Backdoors, DoS, Exploits, Generic, Reconnaissance, Shellcode and Worms. The authors \cite{Moustafa04042016} have generated 49 features from the initial logs plus the class label.
A partition from this dataset is configured as a training set and testing set. In this section we used this given partition. Note: When downloading the dataset from Kaggle we had only 43 features.

\subsubsection{Learning} The process is very close to the one described in Section \ref{sec:training}. However, preliminary results (not presented here) show a significant covariate shift between the training and test sets. Indeed using the same methodology as in \cite{BonduEtAlIJCNN11}, we discover that the ‘id’ variable is the main cause of this drift and the consequences could results in a shift between train and test results. The interested reader may find  on the GitHub page \url{https://github.com/vincentlemaire-labs/CAID2025} the code to detect the drift between train and test dataset and the one to train the classifiers. A good idea could be to conduct an analysis to remove  all the variables that carry the drift as in  \cite{BoulleIJCRS15}, but here for simplicity, and comparison purpose to past papers published on this dataset, only the 'id' variable has been removed.

\subsubsection{Results} We present in Table~\ref{table:results-UNSW} the results obtained with Khiops (without decision trees) as well as those obtained using other classifiers, namely Catboost (CB) \cite{cb}  and Random Forest (RF) \cite{rf}, both with their default parameters in scikit-learn. Note: Khiops is able to handle the UNSW-NB15 dataset directly, as it can handle categorical and numerical variables. However, for RF and CB we had to preprocess the categorical variables using ordinal encoding. We also report in Table~\ref{table:results-UNSW-nrj} the energy consumption in Watt of the three classifiers for their training process, measured using the code carbon library \cite{benoit_courty_2024_11171501} in order to evaluate their energy efficiency \cite{10.1145/3748239.3748247}.

Looking at the results of tables \ref{table:results-UNSW} and \ref{table:results-UNSW-nrj}, we observe several points: the 3 classifiers perform equally well in testing, but Khiops is the most robust (test/train ratio), the least energy-consuming (by a large margin) and the more parsimonious (fewer variables used) which is often desirable to facilitate interpretation. 

\begin{table}[!ht]
    \centering
    %\fontsize{8pt}{8pt}
    \selectfont
    \begin{tabular}{|l|c|c|c|c|}
    \hline	
		 &  \multicolumn{4}{c|}{Accuracy}          \\ \hline
         & Train & Test   & ratio Test/Train  &  \#variables \\\hline
Khiops	 & 0,9225	& 0,9017	&  0,9774     &  15 \\\hline
Random Forest	     & 0,9999	& 0,9005	&  0,9006  &  42     \\\hline
CatBoost & 0,9871	& 0,9021	&  0,9139       &   41 \\\hline
    \end{tabular}
    \vspace{0.2cm}
    \caption{Performances of the tree classifiers.}
    \label{table:results-UNSW}
\end{table}

\begin{table}[!ht]
    \centering
    %\fontsize{8pt}{8pt}
    \selectfont
    \begin{tabular}{|l|c|c|}
    \hline	
		 &  \multicolumn{2}{c|}{Energy to train the classifier} \\ \hline
         &  Energy (W)       & ratio Khiops / Competitor \\\hline
Khiops	 & 2,99  $10^{-4}$ & -   \\\hline
Random Forest	     &  88,73 $10^{-4}$ & 30  \\\hline
CatBoost &   93,24 $10^{-4}$ & 31  \\\hline
    \end{tabular}
    \vspace{0.2cm}
    \caption{Energy Consumption of the tree classifiers}
    \label{table:results-UNSW-nrj}
\end{table}

\subsubsection{Variable importance results} The Figure \ref{fig_visu_imp} shows the normalized importance of the variables for each classifier. There are some similarities (example 'ct\_srv\_dst') but also some marked differences (example 'sttl'). Remember that khiops, being parsimonious, only uses 15 variables. We also give in Table \ref{table:results-5-imp} the first five more important variables for each classifier.

\begin{figure*}[!htb]
\begin{center}
 \includegraphics[width=1.0\textwidth]{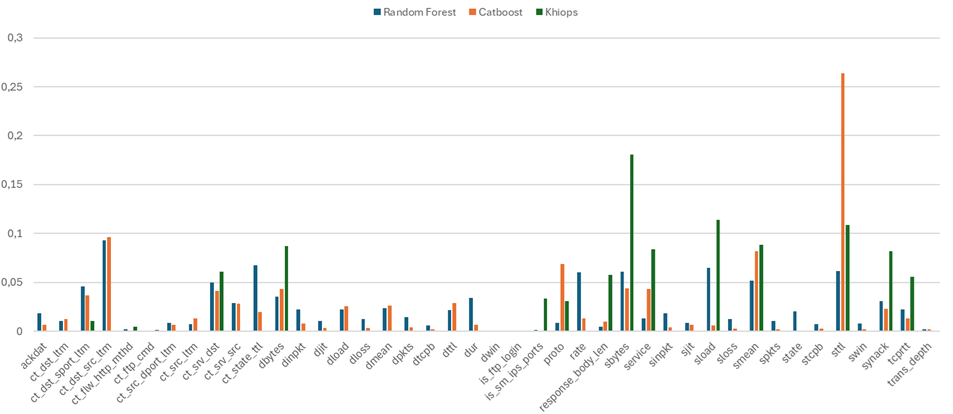}
 \caption{Variable's Importance for the 3 classifieurs}
 \label{fig_visu_imp}
\end{center}
\end{figure*}

\begin{table}[!ht]
    \centering
    %\fontsize{8pt}{8pt}
    \selectfont
    \begin{tabular}{|l|l|l|}
    \hline	
Khiops	&Random Forest	&CatBoost\\\hline
sbytes	&ct\_dst\_src\_ltm	&sttl\\\hline
sload	&ct\_state\_ttl	&ct\_dst\_src\_ltm\\\hline
sttl	&sload	&smean\\\hline
smean	&sttl	&proto\\\hline
dbytes	&sbytes	&sbytes\\\hline
    \end{tabular}
    \vspace{0.2cm}
    \caption{The first five more important variables for each classifier}
    \label{table:results-5-imp}
\end{table}

\vspace{2cm}

%%%%%%%%%%%%%%%%%%%%%%%%%%%%%%%%%%%%%%%%%
%\subsection{\textcolor{magenta}{The CIC-IDS-2017 dataset}}

%\clearpage
%%%%%%%%%%%%%%%%%%%%%%%%%%%%%%%%%%%%%%%%%
\section{Frugal use of computer resources}
\label{zeta_example}

In this section, we illustrate how Khiops makes efficient use of computer resources, enabling the tool to analyze datasets that are much larger than the available RAM.

For this experiment, we use the \emph{Zeta} dataset from the Large Scale Learning Challenge \footnote{\url{https://k4all.org/project/large-scale-learning-challenge/}}, which contains $500,000$ training examples and $2000$
numerical explicative variables. This is a binary classification problem. This data file takes 9.3 GB on the hard disk, and this run was carried out on a Intel Xeon Gold 6150 CPU 2.70 Ghz.

To illustrate the size of the problem, loading the dataset into memory using Python pandas takes about two minutes and requires approximately 8 GB of RAM. Using an optimized \emph{parquet} data format reduces the loading time by about a factor of 15, but the memory footprint remains the same, not accounting for any additional algorithmic requirements for training a classification model.
Analyzing such a dataset is therefore impossible if the available RAM is not significantly larger than the data size.

\begin{figure}[!h]
\begin{center}
 \includegraphics[width=1.0\linewidth]{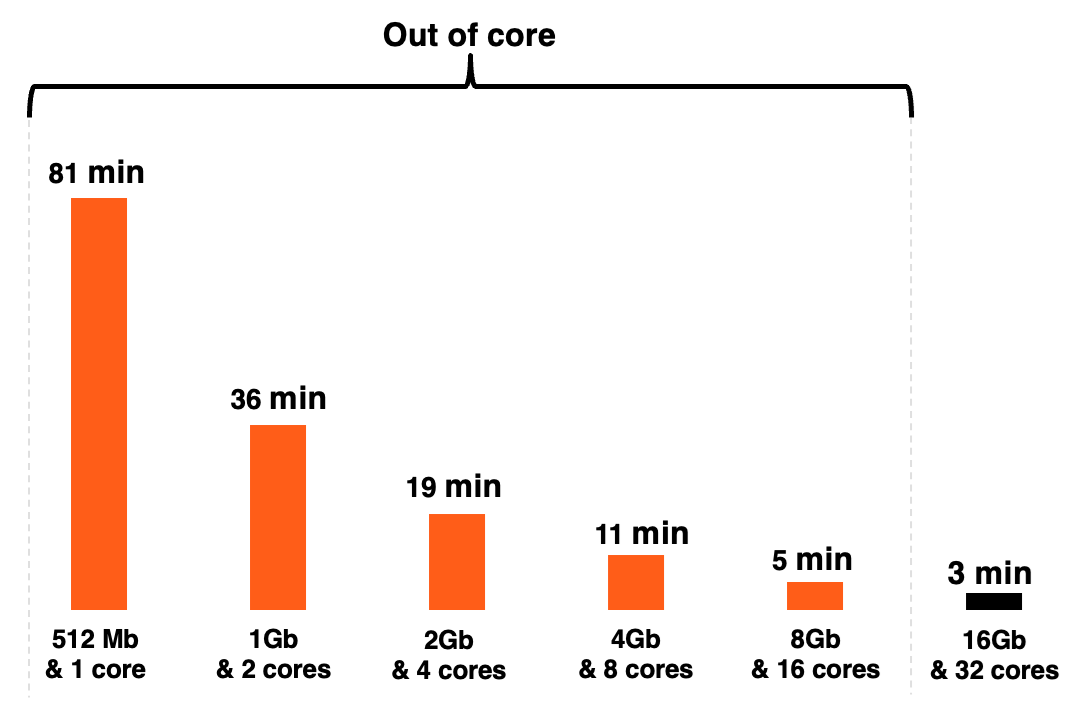}
 \caption{Calculation time for 9 GB dataset.}
 \label{fig_bench_hardware_short}
\end{center}
\end{figure}

Using Khiops, the experiment consists in training a classifier and evaluating it, by varying the number of cores and the amount of RAM available. $70\%$ of the examples are used for training and $30\%$ for testing.
Figure~\ref{fig_bench_hardware_short} plots the execution time in minutes, as the number of cores and the amount of RAM increase together.
Firstly, the results indicate that Khiops can analyze this large dataset using just 512 MB of RAM and a single core.
Due to the limited computational resources, the full processing pipeline takes 81 minutes, whereas it only takes 3 minutes with 32 cores and 16 GB of RAM.
Figure~\ref{fig_bench_hardware_short} shows that there is a smooth transition from out-of-core to distributed computing, demonstrating the efficiency of the adaptation strategy to the available hardware resources. This is made possible by thorough I/O optimization. Finally, you won't be penalized significantly if your hardware is undersized for the task at hand

\section{Perspectives}

Within Orange, major research work is continuing around Khiops, with the release in 2025 of advanced methodologies, such as: robust calibration of classifiers, selection of columns in secondary tables, selection of variables in the presence of concept drift. In the medium term, work will be carried out to process signal-type data (i.e. time series and images) and to develop generative models dedicated to tabular data.   More broadly, the MODL approach has been and continues to be studied by the scientific community, with work on association rules \cite{gay2012bayesian}, sequence mining \cite{egho2017user}, clustering \cite{guigoures2013utilisation,ismaili2016clustering} , uplift \cite{rafla2023bayesian} and multi-table variable selection \cite{boulle2014towards}, for example.

\bibliographystyle{IEEEtran}
\bibliography{References}

\begin{thebibliography}{10}
\providecommand{\url}[1]{#1}
\csname url@samestyle\endcsname
\providecommand{\newblock}{\relax}
\providecommand{\bibinfo}[2]{#2}
\providecommand{\BIBentrySTDinterwordspacing}{\spaceskip=0pt\relax}
\providecommand{\BIBentryALTinterwordstretchfactor}{4}
\providecommand{\BIBentryALTinterwordspacing}{\spaceskip=\fontdimen2\font plus
\BIBentryALTinterwordstretchfactor\fontdimen3\font minus \fontdimen4\font\relax}
\providecommand{\BIBforeignlanguage}[2]{{%
\expandafter\ifx\csname l@#1\endcsname\relax
\typeout{** WARNING: IEEEtran.bst: No hyphenation pattern has been}%
\typeout{** loaded for the language `#1'. Using the pattern for}%
\typeout{** the default language instead.}%
\else
\language=\csname l@#1\endcsname
\fi
#2}}
\providecommand{\BIBdecl}{\relax}
\BIBdecl

\bibitem{ehis2023optimization}
A.-M.~T. Ehis, ``Optimization of security information and event management (siem) infrastructures, and events correlation/regression analysis for optimal cyber security posture,'' \emph{Archives of Advanced Engineering Science}, pp. 1--10, 2023.

\bibitem{zulfadhilah2016cyber}
M.~Zulfadhilah, Y.~Prayudi, and I.~Riadi, ``Cyber profiling using log analysis and k-means clustering,'' \emph{International Journal of Advanced Computer Science and Applications}, vol.~7, no.~7, pp. 430--435, 2016.

\bibitem{boulle:pastel-00003023}
M.~Boull{\'e}, ``{Recherche d'une repr{\'e}sentation des donn{\'e}es efficace pour la fouille des grandes bases de donn{\'e}es},'' Ph.D. dissertation, {T{\'e}l{\'e}com ParisTech}, 2007.

\bibitem{BoulleML06}
M.~Boull\'e, ``{MODL}: a {B}ayes optimal discretization method for continuous attributes,'' \emph{Machine {L}earning}, vol.~65, no.~1, pp. 131--165, 2006.

\bibitem{BoulleJMLR05}
------, ``A {B}ayes optimal approach for partitioning the values of categorical attributes,'' \emph{Journal of {M}achine {L}earning {R}esearch}, vol.~6, pp. 1431--1452, 2005.

\bibitem{voisineEtAlAKDM09}
N.~Voisine, M.~Boull\'e, and C.~Hue, ``A bayes evaluation criterion for decision trees,'' \emph{Advances in Knowledge Discovery and Management (AKDM09)}, vol. 292, pp. 21--38, 2009.

\bibitem{boulle2019scalable}
M.~Boull{\'e}, C.~Charnay, and N.~Lachiche, ``A scalable robust and automatic propositionalization approach for bayesian classification of large mixed numerical and categorical data,'' \emph{Machine Learning}, vol. 108, pp. 229--266, 2019.

\bibitem{hue2024fractionalnaivebayesfnb}
\BIBentryALTinterwordspacing
C.~Hue and M.~Boullé, ``Fractional naive bayes (fnb): non-convex optimization for a parsimonious weighted selective naive bayes classifier,'' 2024. [Online]. Available: \url{https://arxiv.org/abs/2409.11100}
\BIBentrySTDinterwordspacing

\bibitem{SonnenburgEtAl08}
S.~Sonnenburg, V.~Franc, E.~Yom-Tov, and M.~Sebag, ``Pascal large scale learning challenge,'' 2008, http://largescale.first.fraunhofer.de/\-about/.

\bibitem{10.1145/3748239.3748247}
\BIBentryALTinterwordspacing
L.~Arga, F.~B\'{e}lorgey, A.~Braud, R.~Carbou, N.~Charbonniaud, C.~Colomes, L.~Delphin-Poulat, D.~Excoffier, C.~Fauch\'{e}, T.~George, F.~Guyard, T.~Hassan, Q.~Lampin, V.~Lemaire, P.~Nodet, P.~Piotrowski, K.~Sapiejewski, E.~Sirvent-Hien, and T.~Tosic, ``Frugal {AI}: Introduction, {C}oncepts, {D}evelopment and {O}pen {Q}uestions,'' \emph{SIGKDD Explor. Newsl.}, vol.~27, no.~1, p. 72–111, Jul. 2025. [Online]. Available: \url{https://doi.org/10.1145/3748239.3748247}
\BIBentrySTDinterwordspacing

\bibitem{10.1007/978-3-031-74630-7_6}
V.~Lemaire, F.~Cl{\'e}rot, and M.~Boull{\'e}, ``An efficient shapley value computation for the naive bayes classifier,'' in \emph{Machine Learning and Principles and Practice of Knowledge Discovery in Databases}, R.~Meo and F.~Silvestri, Eds.\hskip 1em plus 0.5em minus 0.4em\relax Cham: Springer Nature Switzerland, 2025, pp. 75--90.

\bibitem{lemaire2024viewing}
V.~Lemaire, N.~Le~Boudec, V.~Guyomard, and F.~Fessant, ``Viewing the process of generating counterfactuals as a source of knowledge: a new approach for explaining classifiers,'' in \emph{International Joint Conference on Neural Networks (IJCNN)}.\hskip 1em plus 0.5em minus 0.4em\relax IEEE, 2024, pp. 1--8.

\bibitem{LemaireKDD2009correlation}
\BIBentryALTinterwordspacing
V.~Lemaire, C.~Hue, and O.~Bernier, ``Correlation explorations in a classification model,'' in \emph{Workshop Data Mining Case Studies and Practice Prize, {KDD} 2009}, 2009. [Online]. Available: \url{https://www.researchgate.net/publication/377921678_Correlation_Explorations_in_a_Classification_Model}
\BIBentrySTDinterwordspacing

\bibitem{NIPS2017_7062}
S.~M. Lundberg and S.-I. Lee, ``A unified approach to interpreting model predictions,'' in \emph{Neural Information Processing Society (NeurIPS)}, 2017.

\bibitem{Moustafa04042016}
N.~Moustafa and J.~Slay, ``The evaluation of network anomaly detection systems: Statistical analysis of the unsw-nb15 data set and the comparison with the kdd99 data set,'' \emph{Information Security Journal: A Global Perspective}, vol.~25, no. 1-3, pp. 18--31, 2016.

\bibitem{BonduEtAlIJCNN11}
A.~Bondu and M.~Boullé, ``A supervised approach for change detection in data streams,'' in \emph{Proceedings of International Joint Conference on Neural Networks}, 2011, pp. 519--526.

\bibitem{BoulleIJCRS15}
M.~Boullé, ``Prediction of methane outbreak in coal mines from historical sensor data under distribution drift,'' in \emph{Rough Sets, Fuzzy Sets, Data Mining, and Granular Computing - 15th International Conference, RSFDGrC 2015}, 2015, pp. 439--451.

\bibitem{cb}
L.~Prokhorenkova, G.~Gusev, A.~Vorobev, A.~V. Dorogush, and A.~Gulin, ``Catboost: unbiased boosting with categorical features,'' in \emph{International Conference on Neural Information Processing Systems}, ser. NIPS'18.\hskip 1em plus 0.5em minus 0.4em\relax Red Hook, NY, USA: Curran Associates Inc., 2018, p. 6639–6649.

\bibitem{rf}
L.~Breiman, ``Random forests,'' vol.~45, no.~1, pp. 5--32.

\bibitem{benoit_courty_2024_11171501}
\BIBentryALTinterwordspacing
B.~Courty, V.~Schmidt, S.~Luccioni, Goyal-Kamal, MarionCoutarel, B.~Feld, J.~Lecourt, LiamConnell, A.~Saboni, Inimaz, supatomic, M.~Léval, L.~Blanche, A.~Cruveiller, ouminasara, F.~Zhao, A.~Joshi, A.~Bogroff, H.~de~Lavoreille, N.~Laskaris, E.~Abati, D.~Blank, Z.~Wang, A.~Catovic, M.~Alencon, M.~Stęchły, C.~Bauer, L.~O.~N. de~Araújo, JPW, and MinervaBooks, ``mlco2/codecarbon: v2.4.1,'' May 2024. [Online]. Available: \url{https://doi.org/10.5281/zenodo.11171501}
\BIBentrySTDinterwordspacing

\bibitem{gay2012bayesian}
D.~Gay and M.~Boull{\'e}, ``A bayesian approach for classification rule mining in quantitative databases,'' in \emph{Joint European Conference on Machine Learning and Knowledge Discovery in Databases}.\hskip 1em plus 0.5em minus 0.4em\relax Springer, 2012, pp. 243--259.

\bibitem{egho2017user}
E.~Egho, D.~Gay, M.~Boull{\'e}, N.~Voisine, and F.~Cl{\'e}rot, ``A user parameter-free approach for mining robust sequential classification rules,'' \emph{Knowledge and Information Systems}, vol.~52, no.~1, pp. 53--81, 2017.

\bibitem{guigoures2013utilisation}
R.~Guigour{\`e}s, ``Utilisation des mod{\`e}les de co-clustering pour l'analyse exploratoire des donn{\'e}es,'' Ph.D. dissertation, Universit{\'e} Panth{\'e}on-Sorbonne-Paris I, 2013.

\bibitem{ismaili2016clustering}
O.~A. Ismaili, ``Clustering pr{\'e}dictif d{\'e}crire et pr{\'e}dire simultan{\'e}ment,'' Ph.D. dissertation, Universit{\'e} Paris Saclay (COmUE), 2016.

\bibitem{rafla2023bayesian}
M.~Rafla, ``A bayesian approach for uplift modeling: application on biased data,'' Ph.D. dissertation, Normandie Universit{\'e}, 2023.

\bibitem{boulle2014towards}
M.~Boull{\'e}, ``Towards automatic feature construction for supervised classification,'' in \emph{Joint European Conference on Machine Learning and Knowledge Discovery in Databases}.\hskip 1em plus 0.5em minus 0.4em\relax Springer, 2014, pp. 181--196.

\end{thebibliography}

\end{document}